\documentclass[a4paper]{article}
\usepackage[utf8]{inputenc}
\usepackage{INTERSPEECH2018}
\usepackage[acronym]{glossaries}
\usepackage{multirow}
\usepackage{makecell}
\usepackage{siunitx}

\newacronym{AM}{AM}{acoustic model}
\newacronym{ASR}{ASR}{automatic speech recognition}
\newacronym{BAN}{BAN}{blind analytic normalization}
\newacronym[plural=BLSTMs, firstplural=bidirectional long short-term memories (BLSTMs)]{BLSTM}{BLSTM}{bidirectional long short-term memory}
\newacronym{DAS}{DAS}{delay and sum}
\newacronym{DFT}{DFT}{discrete Fourier transform}
\newacronym{FS}{FS}{filter and sum}
\newacronym[plural=GMMs, firstplural=Gaussian mixture models (GMMs)]{GMM}{GMM}{Gaussian mixture model}
\newacronym[plural=DNNs, firstplural=deep neural networks (DNNs)]{DNN}{DNN}{deep neural network}
\newacronym{GEV}{GEV}{generalized eigenvalue}
\newacronym[plural=HMMs, firstplural=hidden Markov models (HMMs)]{HMM}{HMM}{hidden Markov model}
\newacronym[plural=LSTMs, firstplural=long short-term memories (LSTMs)]{LSTM}{LSTM}{long short-term memory}
\newacronym{MVDR}{MVDR}{minimum variance distortionless response}
\newacronym[plural=RNNs, firstplural=recurrent neural networks (RNNs)]{RNN}{RNN}{recurrent neural network}
\newacronym{SNR}{SNR}{signal-to-noise ratio}
\newacronym{STFT}{STFT}{short-time Fourier transform}
\newcommand{\beamformitInitialAmRescoringDev}{4.36}
\newcommand{\gevInitialAmRescoringDev}{3.46}
\newcommand{\gevJointAmRescoringDev}{3.32}
\newcommand{\gevJointAmSaMeRescoringDev}{3.09}

\newcommand{\gevJointAmRescoringDevFeOne}{4.19}
\newcommand{\gevJointAmSaMeRescoringDevFeOne}{3.55}

\newcommand{\gevJointAmRescoringDevFeFour}{3.23}
\newcommand{\gevJointAmSaMeRescoringDevFeFour}{3.20}

\newcommand{\gevJointAmRescoringDevMaThree}{2.77}
\newcommand{\gevJointAmSaMeRescoringDevMaThree}{2.48}

\newcommand{\gevJointAmRescoringDevMaFour}{3.07}
\newcommand{\gevJointAmSaMeRescoringDevMaFour}{3.14}
\newcommand{\beamformitInitialAmRescoringEval}{7.17}
\newcommand{\gevInitialAmRescoringEval}{5.18}
\newcommand{\gevJointAmRescoringEval}{4.84}
\newcommand{\gevJointAmSaMeRescoringEval}{4.58}

\newcommand{\gevJointAmRescoringEvalFeFive}{6.88}
\newcommand{\gevJointAmSaMeRescoringEvalFeFive}{6.35}

\newcommand{\gevJointAmRescoringEvalFeSix}{4.09}
\newcommand{\gevJointAmSaMeRescoringEvalFeSix}{4.09}

\newcommand{\gevJointAmRescoringEvalMaFive}{3.83}
\newcommand{\gevJointAmSaMeRescoringEvalMaFive}{3.38}

\newcommand{\gevJointAmRescoringEvalMaSix}{4.58}
\newcommand{\gevJointAmSaMeRescoringEvalMaSix}{4.48}

\title{Speaker Adapted Beamforming for Multi-Channel\\ Automatic Speech Recognition}
\name{Tobias Menne, Ralf Schl\"{u}ter, Hermann Ney}
\address{
  Human Language Technology and Pattern Recognition, Computer Science Department, \\RWTH Aachen University, Aachen, Germany
  }
\email{\{menne, schlueter, ney\}@cs.rwth-aachen.de}

\DeclareMathOperator*{\argmax}{argmax}

\begin{document}

\maketitle
\begin{abstract}

This paper presents, in the context of multi-channel ASR, a method to adapt a mask based, statistically optimal beamforming approach to a speaker of interest.
The beamforming vector of the statistically optimal beamformer is computed by utilizing speech and noise masks, which are estimated by a neural network.
The proposed adaptation approach is based on the integration of the beamformer, which includes the mask estimation network, and the acoustic model of the ASR system.
This allows for the propagation of the training error, from the acoustic modeling cost function, all the way through the beamforming operation and through the mask estimation network.
By using the results of a first pass recognition and by keeping all other parameters fixed, the mask estimation network can therefore be fine tuned by retraining. Utterances of a speaker of interest can thus be used in a two pass approach, to optimize the beamforming for the speech characteristics of that specific speaker.
It is shown that this approach improves the ASR performance of a state-of-the-art multi-channel ASR system on the CHiME-4 data. Furthermore the effect of the adaptation on the estimated speech masks is discussed.

\end{abstract}
\noindent\textbf{Index Terms}: robust ASR, multi-channel ASR, speaker adaptation, acoustic beamforming, CHiME-4

\section{Introduction}

The performance of \gls{ASR} systems has shown significant improvements over the last decade. Those have especially been driven by the utilization of deep learning techniques \cite{deeNeuNetForAcoModInSpeRec}.
Nevertheless the performance of systems dealing with realistic noisy and far-field scenarios is still significantly worse than the performance of close talking systems on clean recordings \cite{anOveOfNoiRobAutSpeRec, strForDisSpeRecInRevEnv}. 
Multi-channel \gls{ASR} systems are often used in those scenarios to improve recognition robustness.
In these systems the effect of noise, reverberation and speech overlap is mitigated by utilizing spatial information through beamforming \cite{micArrSigProTecAndApp}.

Usually beamforming is done in a separate preprocessing step before applying the \gls{ASR} system to the enhanced signal, which is obtained from the output of the preprocessing \cite{theRwtUpbForSysComForThe4thChiChaEva}. 
A general formulation for beamforming is the filter-and-sum approach \cite{acoFilAndSumBeaByAdaPriComAna, beaAVerAppToSpatFil}, where the single channels are summed up after applying a separate linear filter to each one. 
Usually those filters are derived such that an objective criterion on the signal level, such as \gls{SNR}, is optimized.
Popular approaches are the \gls{DAS} \cite{micArrSigProTecAndApp}, \gls{MVDR} \cite{impMvdBeaUsiSinChaMasPreNet} and \gls{GEV} \cite{bliAcoBeaBasOnGenEigDec} beamforming methods.
Most systems submitted to the CHiME and REVERB challenges \cite{theThiChiSpeSepAndRecChaDatTasAndBas, anAnaOfEnvMicAndDatSimMisInRobSpeRec, theRevChaAComEvaFraForDerAndRecOfRevSpe} follow one or more of these approaches.

The objective used to optimize the preprocessing thus differs from the objective of the acoustic model training. 
Even before the introduction of \gls{DNN} hybrid systems in \gls{ASR}, the optimization of the preprocessing towards the goal of speech recognition was proposed e.g. in \cite{likMaxBeaForRobHanSpeRec}. 
The success of deep learning also motivated the integration of the beamforming operation into the acoustic model. 
E.g. in \cite{deeBeaNetForMulChaSpeRec, deeLonShoTerMemAdaBeaNetForMulRobSpeRec} the filters of the filter-and-sum beamforming are estimated by a neural network based on input features derived from the multi-channel input signal.
Even learning the complete multi-channel preprocessing, starting from the raw time signal, has been shown to work \cite{speLocAndMicSpaInvAcoModFroRawMulWav, facSpaAndSpeMulRawWavCld, neuNetAdaBeaForRobMulSpeRec}.
The advantage of those approaches is, that the preprocessing is not optimized for a proxy measure like \gls{SNR} at the output of the beamformer, but directly towards the criterion for acoustic model training. 
But thus far, a very large amount of training data is necessary to obtain satisfying performance with those approaches.

Lately the performance of statistically optimal beamformers was improved by using neural networks to estimate speech and noise masks, which are then used to compute the beamforming vectors \cite{blsSupGevBeaFroForThe3rdChiChal, robMvdBeaUsiTimFreMasForOnlOffAsrInNoi, impMvdBeaUsiSinChaMasPreNet}.
This approach has worked well for many submissions to the $\mathrm{4}^{\mathrm{th}}$ CHiME challenge \cite{theRwtUpbForSysComForThe4thChiChaEva, widResBlsNetWitDisSpeAdaForRobSpeRec, theUstIflSysForChiCha}.
One problem of that approach is the need for target masks in the mask estimator training, which usually requires stereo data (the noisy and its respective clean signal) to create the target masks for training. 
Since this type of data is much more difficult to collect than only the noisy data, training of the mask estimator is usually done on simulated signals, which can lead to a mismatch between training and test data.
To solve this problem, the authors of \cite{beaEndToEndTraOfABeaSupMulChaAsrSys} proposed to integrate the mask based, statistically optimal beamforming with the acoustic modeling of the \gls{ASR} system. 
This enables the propagation of the training error all the way through the acoustic model and the mask estimator network in the preprocessing.
Therefore the mask estimator can be trained based on the training criterion of the acoustic model training.

In this paper, the approach of integrating the mask based, statistically optimal beamformer with the acoustic model is utilized to adapt the mask estimation to the speech characteristics of a speaker of interest in a two pass recognition approach.

The rest of the paper is organized as follows. 
An overview of the integrated system is given in Section \ref{sec_systemOverview}. Furthermore an alternative approach to \cite{beaEndToEndTraOfABeaSupMulChaAsrSys} for the propagation of the gradients through the eigenvalue problem of the beamformer is presented. 
Section \ref{sec_experimentalSetup} describes the experimental setup of a state-of-the-art system for the CHiME-4 speech recognition task 
followed by the experimental results in Section \ref{sec_experimentalResults}.

\section{System overview}
\label{sec_systemOverview}

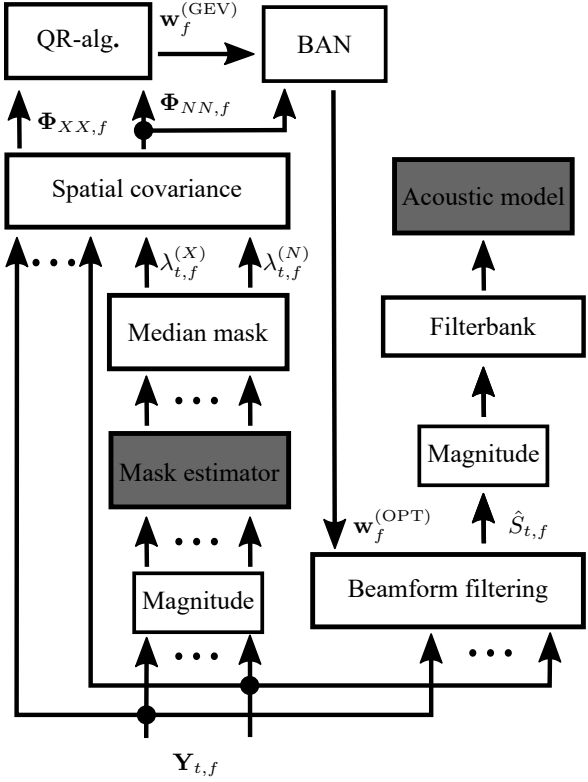
\begin{figure}[t]
\setlength{\unitlength}{\textwidth}
\begin{picture}(0.5, 0.66)
\put(0, 0.025){\includegraphics[width=0.45\textwidth]{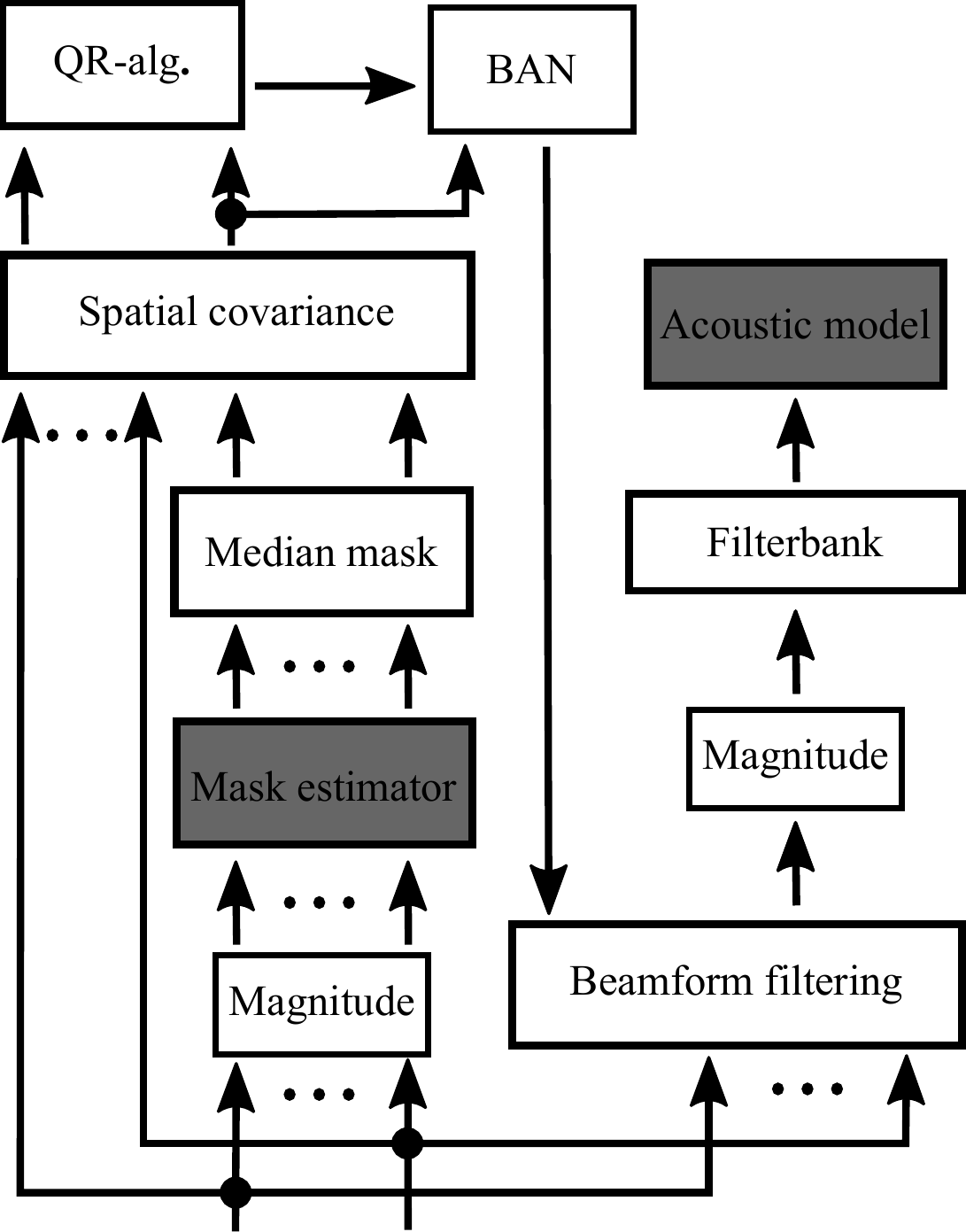}}
\put(0.13, 0){\normalsize $\mathbf{Y}_{t, f}$}
\put(0.27, 0.185){\normalsize $\mathbf{w}^{\mathrm{(OPT)}}_{f}$}
\put(0.12, 0.515){\normalsize $\mathbf{\Phi}_{NN, f}$}
\put(0.025, 0.5){\normalsize $\mathbf{\Phi}_{XX, f}$}
\put(0.12, 0.58){\normalsize $\mathbf{w}^{\mathrm{(GEV)}}_{f}$}
\put(0.39, 0.185){\normalsize $\hat{S}_{t, f}$}
\put(0.12, 0.39){\normalsize $\lambda^{\mathrm{(}X\mathrm{)}}_{t, f}$}
\put(0.2, 0.39){\normalsize $\lambda^{\mathrm{(}N\mathrm{)}}_{t, f}$}
\end{picture}
  \caption{Overview of the integrated system. The grey blocks indicate modules with trainable parameters.}
  \label{fig_pipeline}
\end{figure}

The system used in this work integrates the acoustic beamformer, usually called front-end, 
with the acoustic model of the \gls{ASR} system, usually called back-end, 
very similarly to the integration described in \cite{beaEndToEndTraOfABeaSupMulChaAsrSys}. 
Figure \ref{fig_pipeline} gives an overview of the integrated system.
$\mathbf{Y}_{t, f}$ is the input in the \gls{STFT} domain, recorded from an array of $M$ microphones. It consists of a speech component $\mathbf{X}_{t, f}$ and a noise component $\mathbf{N}_{t, f}$:
\begin{equation}
\mathbf{Y}_{t, f} = \mathbf{X}_{t, f} + \mathbf{N}_{t, f}
\end{equation}
Where $\mathbf{Y}_{t, f}, \mathbf{X}_{t, f}, \mathbf{N}_{t, f} \in \mathbb{C}^{M}$, $t$ is the time frame index and $f$ is the frequency bin index.

The main difference to the system introduced in \cite{beaEndToEndTraOfABeaSupMulChaAsrSys} will be described in Section \ref{sec_systemOverview_beamformerIntegration},
whereas the acoustic beamformer and acoustic model are described in Sections \ref{sec_systemOverview_beamformer} and \ref{sec_systemOverview_acousticModel}, respectively. 
During a first pass decoding a \gls{HMM}-state sequence $s_{1}^{T}$ is obtained for the input signal $\mathbf{Y}_{1, 1}^{T, F}$, where $T$ and $F$ are the number of time frames and frequency bins of the signal. 
Section \ref{sec_systemOverview_adaptation} describes the utilization of the state sequence to adapt the acoustic beamformer to a certain speaker.

\subsection{{GEV} beamformer}
\label{sec_systemOverview_beamformer}

The main purpose of the front-end is to denoise the input signal. 
Here this is achieved by acoustic beamforming \cite{acoFilAndSumBeaByAdaPriComAna, beaAVerAppToSpatFil}:
\begin{equation}
\hat{S}_{t, f} = \mathbf{w}^{H}_{f} \cdot \mathbf{Y}_{t, f}
\end{equation}
Where $\hat{S}_{t, f} \in \mathbb{C}$ is an estimate of the speech component, obtained by applying the beamforming vector $\mathbf{w}_{f} \in \mathbb{C}^{M}$. $(\cdot)^{H}$ denotes the Hermitian transpose.

For this work we use the \gls{GEV} beamformer with \gls{BAN}, as described in \cite{bliAcoBeaBasOnGenEigDec} and which is also used in \cite{beaEndToEndTraOfABeaSupMulChaAsrSys}.
The beamforming vector of the \gls{GEV} beamformer is derived by maximizing the {\it a posteriori} \gls{SNR}:
\begin{equation}
\mathbf{w}^{\mathrm{(GEV)}}_{f} = \argmax_{\mathbf{w}_{f}}\frac{\mathbf{w}^{H}_{f}\mathbf{\Phi}_{XX, f}\mathbf{w}_{f}}{\mathbf{w}^{H}_{f}\mathbf{\Phi}_{NN, f}\mathbf{w}_{f}}
\end{equation}
Where $\mathbf{\Phi}_{XX, f}$ and $\mathbf{\Phi}_{NN, f}$ are the spatial covariance matrices of speech and noise, respectively.
This results in the generalized eigenvalue problem
\begin{equation}
\label{eq_genEigValProblem}
\mathbf{\Phi}_{XX, f}\mathbf{W} = \mathbf{\Phi}_{NN, f}\mathbf{W}\mathbf{\Lambda}
\end{equation}
with $\mathbf{w}^{\mathrm{(GEV)}}_{f}$ being the eigenvector corresponding to the largest eigenvalue.

The spatial covariance matrices $\mathbf{\Phi}_{\nu\nu, f}$ for $\nu \in \lbrace X, N \rbrace$ are computed by applying a mask $\lambda^{\mathrm{(}\nu\mathrm{)}}_{t, f}$ to the recorded multi-channel signal $\mathbf{Y}_{t, f}$
\begin{equation}
\mathbf{\Phi}_{\nu\nu, f} = \frac{1}{\sum_{t=1}^{T}{\lambda^{\mathrm{(}\nu\mathrm{)}}_{t, f}}}\sum_{t=1}^{T}{\lambda^{\mathrm{(}\nu\mathrm{)}}_{t, f} \mathbf{Y}_{t, f} \mathbf{Y}_{t, f}^{H}}
\end{equation}
A mask estimating neural network is used to estimate $\lambda^{\mathrm{(}X\mathrm{)}}_{t, f}$ and $\lambda^{\mathrm{(}N\mathrm{)}}_{t, f}$.
For both, speech and noise, one mask is estimated for every channel, $\lambda^{\mathrm{(}\nu\mathrm{)}}_{t, f}$ is then computed as the median mask, which contains the element-wise median of the channel dependent masks, as described e.g. in \cite{blsSupGevBeaFroForThe3rdChiChal}. 

The \gls{BAN} post-filter, as described in \cite{bliAcoBeaBasOnGenEigDec}, is a frequency dependent scaling of the \gls{GEV} beamforming vector, such that the final beamforming vector used here is:
\begin{equation}
\mathbf{w}^{\mathrm{(OPT)}}_{f} = w^{\mathrm{(BAN)}}_{f} \cdot \mathbf{w}^{\mathrm{(GEV)}}_{f}
\end{equation}
With $w^{\mathrm{(BAN)}}_{f} \in \mathbb{C}$ being the scaling factor described in \cite{bliAcoBeaBasOnGenEigDec}.

\subsection{Acoustic model}
\label{sec_systemOverview_acousticModel}
The acoustic model is a \gls{BLSTM} hybrid model using log-mel filterbank features as input.
Apart from the features, the training pipeline is the same as for the speaker independent model described in \cite{theRwtUpbForSysComForThe4thChiChaEva}.

\subsection{Beamformer integration into acoustic model}
\label{sec_systemOverview_beamformerIntegration}

Training of the integrated system presented in Figure \ref{fig_pipeline} is done according to standard error back propagation. 
The gradient computation for the propagation through the acoustic model, feature extraction, linear filtering of the beamformer, \gls{BAN} and mask estimator network are straight forward.
To propagate the gradient through the computation of the principal eigenvector of 
\begin{equation}
\label{eq_eigVecPhi}
\mathbf{\Phi}_{f} = \mathbf{\Phi}_{NN, f}^{-1}\mathbf{\Phi}_{XX, f}
\end{equation}
as required for computing the beamforming vector $\mathbf{w}^{\mathrm{(GEV)}}_{f}$ according to Equation \ref{eq_genEigValProblem},
the derivatives of the eigenvalue problem w.r.t. $\mathbf{\Phi}_{NN, f}$ and $\mathbf{\Phi}_{XX, f}$ are derived in \cite{optNeuNetSupAcoBeaByAlgDif} and used in \cite{beaEndToEndTraOfABeaSupMulChaAsrSys}. 

In contrast, here the principal eigenvector of Equation \ref{eq_eigVecPhi} is approximated by applying the QR-algorithm as presented in \cite{theQrTraAUniAnaToTheLrTraPar1}. 
A matrix $A_{k}$ is decomposed by the QR-decomposition into a product of a unitary matrix $Q_{k}$ and an upper triangular matrix $R_{k}$:
\begin{equation}
A_{k} = Q_{k} R_{k}
\end{equation}
With $k$ being the iteration index, $A_{k+1}$ is then computed as
\begin{equation}
A_{k+1} = R_{k} Q_{k}
\end{equation}
It is shown in \cite{theQrTraAUniAnaToTheLrTraPar1}, that $A_{K}$ converges to an upper triangular matrix as $K \rightarrow \infty$.
The diagonal of $A_{K}$ then contains the eigenvalues of $A_{0}$ and $\prod_{k=0}^{K} Q_{k}$ contains the respective eigenvectors. 
This QR-algorithm is used here to approximate the principal eigenvector of $\mathbf{\Phi}_{f}$ by setting
\begin{equation}
\label{eq_qrInitialization}
A_{0} = \mathbf{\Phi}_{f}
\end{equation}

The algorithmic differentiation of the QR decomposition is outlined in \cite{onEvaHigOrdDerOfTheQrDecOfTalMatWitFulColRanInForAndRevModAlgDif} and applied here in the error back propagation.

\subsection{Speaker adaptation of mask estimator}
\label{sec_systemOverview_adaptation}

After a first pass recognition an optimal sequence of HMM states $s_{t}^{T}$ is obtained from the decoding process for each of the evaluation segments of the speaker of interest. 
Those alignments are then used as training targets for an adaptation training of the integrated system. 
Of the system shown in Figure \ref{fig_pipeline}, only the parameters of the mask estimator are adjusted in the adaptation training. 
The parameters of the remaining pipeline are kept fixed, such that only the mask estimator network is tuned towards optimizing the cost function of the integrated system. 
Therefore the mask estimator and thus the computation of the beamforming vector are optimized for the speech characteristics of the speaker of interest.

Even though this work is using the \gls{GEV} beamformer with \gls{BAN}, it is noteworthy that the proposed speaker adaptation method is equally applicable to the mask based \gls{MVDR} beamformer that is presented in \cite{robMvdBeaUsiTimFreMasForOnlOffAsrInNoi}, by changing the initialization of $A_{0}$ in Equation \ref{eq_qrInitialization} and omitting the \gls{BAN}.

\section{Experimental setup}
\label{sec_experimentalSetup}

The proposed speaker adaptation scheme for the acoustic beamformer is evaluated on the data of the CHiME-4 speech recognition task \cite{anAnaOfEnvMicAndDatSimMisInRobSpeRec}.
The CHiME-4 dataset features real and simulated \SI{16}{kHz}, multi-channel audio data recorded with a six channel microphone array arranged around a tablet device.
Based on the 5k WSJ0-Corpus recordings and simulations have been done with four different kinds of real-world background noise.
The training set contains approximately \SI{18}{h} of data per channel recorded from 87 different speakers.
Results are provided for the real development and real evaluation set of the 6-Channel track.
Both sets contain audio of 4 speakers each, of which 2 are male and 2 are female, with no overlap between development and evaluation set.
The amount of data per speaker is approximately \SI{0.7}{h} in the development set and around \SI{0.5}{h} in the evaluation set.

The acoustic model used in the experiments is a \gls{BLSTM} network, with 5 layers and 600 units per layer.
Different to the system in \cite{theRwtUpbForSysComForThe4thChiChaEva}, the input features are 80 dimensional log-mel filterbank features computed in the \gls{STFT} domain employing a blackman window with a window size of \SI{25}{ms} and a frame shift of \SI{10}{ms}.
The input features are unnormalized, but a linear layer with 80 units, employing batch normalization, was added as a first layer to the network.
This results in a marginally better baseline system over the one described in \cite{theRwtUpbForSysComForThe4thChiChaEva}.
The initial training of the acoustic model is done as described in \cite{theRwtUpbForSysComForThe4thChiChaEva}, 
where at first alignments for the training set are computed on the data of the close talking microphone by using a \gls{GMM}-\gls{HMM} trained only on the data of the close talking microphones of the training set. 
Those alignments can then be used for all other channels, since the data is recorded sample synchronized.
The training of the \gls{BLSTM} acoustic model is done by using the unprocessed audio data of the single channels. 
This has been demonstrated to be beneficial in many submissions to the $\mathrm{3}^{\mathrm{rd}}$ and $\mathrm{4}^{\mathrm{th}}$ CHiME challenge e.g. in \cite{theNttChiSysAdvInSpeEnhAndRecForMobMulMicDev}.

The mask estimator network used in the experiments is similar to the one described in \cite{blsSupGevBeaFroForThe3rdChiChal}.
It consists of a \gls{BLSTM} layer with 256 units followed by two fully connected layers with 512 units and ReLU activations and another fully connected layer with 402 units and sigmoid activation. 
Thus the resolution of the estimated masks in frequency is lower than described in \cite{blsSupGevBeaFroForThe3rdChiChal}. 
This is due to the adjustment of the dimensions of the masks to the \gls{DFT} size of the feature extraction pipeline of the \gls{ASR} system used here.
The input of the mask estimation network is the magnitude spectrum of a single channel.
The output of the network is the concatenation of the noise mask and the speech mask.
During decoding the outputs of the different channels, of one utterance, are grouped and the median masks are calculated. Those are then applied to all channels to estimate the spatial covariance matrices as described in Section \ref{sec_systemOverview_beamformer}.
The initial mask estimation network is trained on the simulated training data as described in \cite{blsSupGevBeaFroForThe3rdChiChal}.
In contrast to \cite{blsSupGevBeaFroForThe3rdChiChal}, only the provided baseline configuration of the simulation is used and no additional data augmentation is done.
The number of iterations of the QR-algorithm described in Section \ref{sec_systemOverview_beamformerIntegration} is fixed to 5.

The decoding is done with the 5-gram language model provided as a baseline language model with the CHiME-4 dataset. 
In a post processing step a \gls{RNN} language model lattice rescoring is done. 
The \gls{RNN} language model is a 3 layer \gls{LSTM} with highway connections. Details about the language model and lattice rescoring can be found in \cite{theRwtUpbForSysComForThe4thChiChaEva}.

In addition to the acoustic beamforming described in Section \ref{sec_systemOverview_beamformer}, the baseline beamforming algorithm of the CHiME-4 task (BFIT) is used to provide baseline results. 
Apart from the beamforming algorithm, the exact same pipeline as described above is used.

The hyper-parameters for the speaker adaptation training such as the learning rate were tuned on the development set and applied to the evaluation set.

\section{Experimental results}
\label{sec_experimentalResults}

\subsection{Baseline systems}

Table \ref{tbl_resultsOverview} shows an overview of the experimental results.
It shows, that using the \gls{GEV} front-end described in Section \ref{sec_systemOverview_beamformer} yields an improvement of about \SI{20}{\percent} - \SI{30}{\percent} relative over the baseline system with the BFIT front-end. Joint training of the \gls{GEV} front-end and acoustic model further improves the performance another \SI{5}{\percent} relative. 
Those results are in line with the results reported in \cite{beaEndToEndTraOfABeaSupMulChaAsrSys}.
When comparing the mask output of the mask estimator before and after joint training only minor differences in the masks can be observed.
This is in line with the suggestion of the authors of \cite{beaEndToEndTraOfABeaSupMulChaAsrSys}, that a majority of the performance increase stems from the adaptation of the acoustic model towards the specific front-end.

\begin{table}[!h] 
\centering 
\caption{Average WER (\%) for the described systems for different stages of the integrated training.} 
\label{tbl_resultsOverview} 
\setlength\tabcolsep{4pt} 
\begin{tabular}{|c|c|c|c|c|c|}     
\hline     
\multicolumn{4}{|c|}{System} & \multirow{2}{*}{Dev} & \multirow{2}{*}{Eval} \\     
\cline{1-4}
\thead{System\\id} & \thead{Front-\\ end} & \thead{Joint \\ training} & \thead{Speaker\\ adapted} & & \\
\hline     
0 & BFIT & \multirow{2}{*}{-} & \multirow{3}{*}{-} & \beamformitInitialAmRescoringDev & \beamformitInitialAmRescoringEval \\
\cline{1-2}
\cline{5-6}
1 & \multirow{3}{*}{GEV} & & &\gevInitialAmRescoringDev & \gevInitialAmRescoringEval\\
\cline{1-1}
\cline{3-3}
\cline{5-6}
2 & & \multirow{2}{*}{+} & &\gevJointAmRescoringDev & \gevJointAmRescoringEval\\
\cline{1-1}
\cline{4-4}
\cline{5-6}
3 & & & + & \bf \gevJointAmSaMeRescoringDev & \bf \gevJointAmSaMeRescoringEval\\
\hline     
\end{tabular}
\end{table}

\subsection{Speaker adapted beamforming}

\begin{figure}[t]
  \centering
  \includegraphics[width=\linewidth]{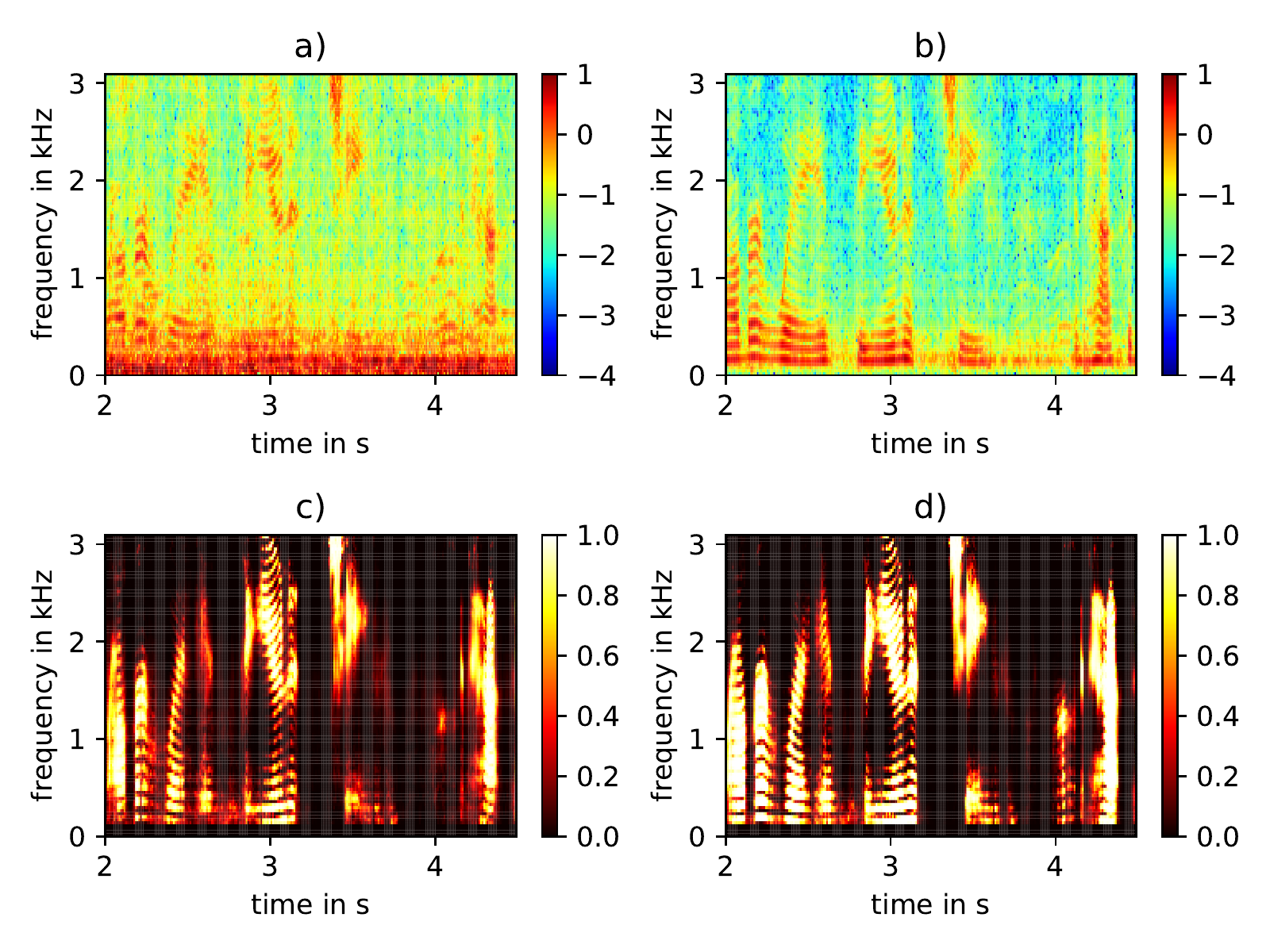}
  \caption{Two seconds snippet of the signal "F01\_421C0210\_BUS" of the development set starting at second 2 and showing the frequency range up to \SI{3}{kHz}. a) log magnitude spectrum of the noisy signal recorded at channel 5 b) log magnitude spectrum of the signal recorded at the close talking microphone c) estimated speech mask of system 2 (jointly trained but before speaker adaptation) d) estimated speech mask of system 3 (after speaker adaptation)}
  \label{fig_exampleMasks}
\end{figure}

Table \ref{tbl_resultsOverview} shows an overall improvement of WER after speaker adaptation and Table \ref{tbl_resultsSpeaker} shows that improved performance is obtained for the majority of the speakers with an improvement in WER of up to \SI{11}{\percent} and \SI{15}{\percent} relative for single speakers of the evaluation and development set, respectively. 
Figure \ref{fig_exampleMasks} shows an example of the estimated speech mask before and after the speaker adaptation.
It can be seen, that the speech mask after speaker adaptation shows a stronger emphasis on the fundamental frequency and the harmonics.
This can be seen repeatedly between the time marks of \SI{2}{s} and \SI{3}{s}. 
At time mark \SI{4}{s} a pattern of fundamental frequency and harmonics can be seen in the mask after adaptation, which is not present in the mask before adaptation and which can also hardly be spotted in the input signal or the clean signal. 
This could indicate an increased bias of the mask estimator towards this kind of pattern.

\begin{table}[!h] 
\centering 
\caption{WER (\%) of separate speakers for the jointly trained system and the speaker adapted system} 
\label{tbl_resultsSpeaker} 
\setlength\tabcolsep{4pt} 
\begin{tabular}{|c|c|c|c|c|c|c|c|c|}     
\hline     
\multirow{2}{*}{\thead{Sys.\\id}} & \multicolumn{4}{|c|}{Dev} & \multicolumn{4}{|c|}{Eval} \\     
\cline{2-9}
 & F01 & F04 & M03 & M04 & F05 & F06 & M05 & M06 \\
\hline     
 2 & \gevJointAmRescoringDevFeOne & \gevJointAmRescoringDevFeFour & \gevJointAmRescoringDevMaThree & \bf \gevJointAmRescoringDevMaFour & \gevJointAmRescoringEvalFeFive & \bf \gevJointAmRescoringEvalFeSix & \gevJointAmRescoringEvalMaFive & \gevJointAmRescoringEvalMaSix \\
\hline     
 3 & \bf \gevJointAmSaMeRescoringDevFeOne & \bf \gevJointAmSaMeRescoringDevFeFour & \bf \gevJointAmSaMeRescoringDevMaThree & \gevJointAmSaMeRescoringDevMaFour & \bf \gevJointAmSaMeRescoringEvalFeFive & \bf \gevJointAmSaMeRescoringEvalFeSix & \bf \gevJointAmSaMeRescoringEvalMaFive & \bf \gevJointAmSaMeRescoringEvalMaSix \\
\hline     
\end{tabular}
\end{table}

\section{Conclusion}

This work describes a method for speaker adaptation of mask based beamforming in a multi-channel ASR system.
The basis of the adaptation method is the integration of the statistically optimal beamforming with the acoustic model to allow the back propagation of the training errors through the complete system, which has been previously introduced in \cite{beaEndToEndTraOfABeaSupMulChaAsrSys}.
Here an alternative solution for the back propagation of the errors through the computation of the beamforming vector, based on the QR-algorithm, is presented.
The system is then used in a two pass approach to adapt the mask estimator to a speaker of interest during the decoding phase.
It was shown that this adaptation method results in speech masks which show a stronger emphasis on the fundamental frequency and harmonics of the speaker.
Furthermore a relative ASR improvement, for single speakers of the real evaluation data of the CHiME-4 ASR task, of up to \SI{11}{\percent} relative was shown.

\section{Acknowledgements}
This project has received funding from the European Research Council (ERC) under the European Union's Horizon 2020 research and innovation program grant agreement No. 694537. 
This work has also been supported by Deutsche Forschungsgemeinschaft (DFG) under contract No. Schl2043/1-1 and European Union's Horizon 2020 research and innovation program under the Marie Sk\l{}odowska-Curie grant agreement No. 644283.
The work reflects only the authors' views and the European Research Council Executive Agency is not responsible for any use that may be made of the information it contains.
The GPU cluster used for the experiments was partially funded by Deutsche Forschungsgemeinschaft (DFG) Grant INST 222/1168-1.

\bibliographystyle{IEEEtran}

\bibliography{mybib}

\end{document}